\documentclass[compsoc]{IEEEtran}
\ifCLASSINFOpdf
   \usepackage[pdftex]{graphicx}
   \graphicspath{{./imgs/}}
  \DeclareGraphicsExtensions{.pdf,.jpeg,.png}
\else
\fi
\usepackage[cmex10]{amsmath}
\usepackage{url}

\usepackage{hyperref}

\begin{document}
%
\title{A Model for Foraging Ants, Controlled by Spiking Neural Networks and Double Pheromones}

\author{\IEEEauthorblockN{Cristian Jimenez-Romero,
David Sousa-Rodrigues
and Jeffrey H. Johnson\\}
\IEEEauthorblockA{Faculty of Maths, Computing and Technology\\
The Open University\\
Milton Keynes\\
MK7 6AA, United Kingdom\\
Emails: cristian.jimenez-romero@open.ac.uk, \\
david@davidrodrigues.org, jeff.johnson@open.ac.uk\\~\\}
\and
\IEEEauthorblockN{Vitorino Ramos\\}
\IEEEauthorblockA{LaSEEB - Evolutionary Systems\\ and Biomedical Engineering Lab.,\\ IST, Technical University of Lisbon,\\ Lisbon, Portugal\\
Email: vitorino.ramos@ist.utl.pt}
}


%


\maketitle

\begin{abstract}
A model of an Ant System where ants are controlled by a spiking neural circuit and a second order pheromone mechanism in a foraging task is presented.  A neural circuit is trained for individual ants and subsequently the ants are exposed to a virtual environment where a swarm of ants performed a resource foraging task. 
The model comprises an associative and unsupervised learning strategy for the neural circuit of the ant. The neural circuit adapts to the environment by means of classical conditioning. The initially unknown environment includes different types of stimuli representing food (rewarding) and obstacles (harmful) which, when they come in direct contact with the ant, elicit a  reflex response in the motor neural system of the ant: moving towards or away from the source of the stimulus. The spiking neural circuits of the ant is trained to identify food and obstacles and move towards the former and avoid the latter.
The ants are released on a landscape with multiple food sources where one ant alone would have difficulty harvesting the landscape to maximum efficiency. In this case the introduction of a double pheromone mechanism (positive and negative reinforcement feedback) yields better results than traditional ant colony optimization strategies.
Traditional ant systems include mainly a positive reinforcement pheromone. This approach uses a second pheromone that acts as a marker for forbidden paths (negative feedback). This blockade is not permanent and is controlled by the evaporation rate of the pheromones. The combined action of both pheromones acts as a collective stigmergic memory of the swarm, which reduces the search space of the problem.
This paper explores how the adaptation and learning abilities observed in biologically inspired cognitive architectures is synergistically enhanced by swarm optimization strategies. The model portraits two forms of artificial intelligent behaviour: at the individual level the spiking neural network is the main controller and at the collective level the pheromone distribution is a map towards the solution emerged by the colony. The presented model is an important pedagogical tool as it is also an easy to use library that allows access to the spiking neural network paradigm from inside Netlogo---a language used mostly in agent based modelling and experimentation with complex systems.
\end{abstract}


%
\IEEEpeerreviewmaketitle

\section{Introduction}

The exploration of artificially constructed entities that simulate biology counterparts has a long tradition in the field of Artificial Intelligence, and in particular in Artificial life. Langton proposed the study of artificial life with celular automata in 1996\,\cite{langton:1986} where the author aimed to ``implement the `molecular logic of the living state' in an artificial biochemistry environment'' via modeling  the artificial molecules as ``virtual automata that were able to roam in an abstract computer space and interact with each other''.

Ant colony based algorithms have been applied successfully to several domains, namely for the clustering of web usage mining\,\cite{Abraham:2003}, image retrieval\,\cite{Ramos:2002a}, newspaper and document organisation\,\cite{Ramos:2002,sousa-rodrigues:2014traversing-news}, and mainly to the travelling salesman problem\,\cite{Bonabeau:1999,Dorigo:1996,Dorigo:1999a,Gambardella:1995,Gupta:2010,Ramos:2013qy}, among other combinatorial optimisation problems.

In nature, many ant species have used trail-laying when foraging. They deposit pheromone, a chemical substance that is volatile and is secreted by the ants when returning to the nest carrying food, that acts as a recruitment mechanism. This pheromone is detected by other colony ants that can use it as an indicator for the food source. This process of ant recruitment is a positive reinforcement mechanism as more ants are driven to the foraging task and subsequently deposit more pheromone. This positive amplification allows the colony to exploit the food source in optimal time. In the model presented here the only process of recruitment is based on chemicals trails and therefore is refereed as \textit{mass recruitment}\,\cite{Bonabeau:1999}.

Social insects are very well adapted to solve the foraging problem. The resilience of the species relies on the flexibility presented by social insects to changing problem landscapes. As sources of food are depleted their strategy needs to allow them to find newer sources and they need to adapt quickly to new situations. The use of pheromones allows them to collectively adapt to changing environments, and the colonies become robust entities even if some individuals fail to perform their tasks. This ``swarm intelligence'' reflects the fact that these social insects are capable of self-organisation. Self-organisation is a process, whereby the dynamics of the parts of the system, a high order structure emerges. Self-organisation systems like ant colonies present defining features: they exhibit some kind of positive feedback mechanism, they exhibit also negative feedback mechanisms, they amplify fluctuations observed in nature and the self-organisation relies on the existence of multiple interactions. Although one single ant can deposit a trail of pheromone, the benefit of these trails is only observed when many ants interact together via the recruitment process.

While ants can interact with other ants in a direct way, via antennation or prophallaxis for example, in this model we only model indirect interaction via the environment. This process is known as stigmergy. One ant deposits pheromones at a certain time and in a certain location and other ants will interact with the deposited pheromone at a latter time. 

Following this introduction, section \ref{sec:model} presents the model in its constituent parts. Subsections \ref{sub:snnvsann} and \ref{sub:snnsuitab} discuss the use of spiking neural networks---compared to traditional artificial neural networks and in the context of controllers for autonomous systems, respectively. Subsection \ref{sub:snn} presents the spiking neural network submodel while subsection \ref{sub:brain} shows the brain of the virtual ant. It is followed in subsection \ref{sub:pheromone} by the description of the double pheromone mechanism and section \ref{sec:model} concludes with details of the implementation in subsection~\ref{sub:netlogo}. Illustrative results are presented in section \ref{sec:results} and it is followed by the conclusion in section \ref{sec:conclusion}. A simplified version of the spiking neural network is available for download at \texttt{\url{http://modelingcommons.org/browse/one_model/4455}}.

\section{A Model for Internal+External Artificial Intelligence\label{sec:model}}

\subsection{Spiking versus traditional Neural Networks\label{sub:snnvsann}}

In traditional Artificial Neural Networks (ANNs) models (e.g. McCulloch-Pitts and sigmoidal neurons) the neural activity is represented by the neurons firing rates, thus the timing between single pulses (Pulse code) is not taken into account \cite{Maass:1998}. On the other hand, Spiking Neural Networks (SNNs) are able to represent neural activity in terms of both rate and pulse codes. The pulse code which makes use of the information contained in the interspike interval has been associated with fast information processing in the brain in cases where the time required for the integration and computation of average rates (rate code) would take too long (i.e., a housefly can change the direction of its flight in reaction to visual stimuli in about 30 milliseconds)\cite{Maass:1998}.
In addition to the enhanced computational capabilities added by the time dimension, the resemblance of SNNs dynamics with their biological counterparts is significantly more accurate than in first and second generations ANNs. SNNs are capable of simulating a broad range of learning and spiking dynamics observed in biological neurons including: Spike timing dependent plasticity (STDP), long term potentiation and depression (LTP/LTD), tonic and phasic spike, inter-spike delay (latency), frequency adaptation, resonance and input accommodation \cite{Izhikevich2003}.  
The STDP rule mentioned above, is implemented in the proposed neural circuit as the underlying mechanism for associative and classical conditioning learning. Experimental results have demonstrated that different types of classical conditioning (i.e., Pavlovian, extinction, partial conditioning, inhibitory conditioning) can be implemented successfully in SNNs. \cite{Liuetal2007}, \cite{Haenicke2012}.

\subsection{Suitability of Spiking neural networks for controlling autonomous systems (robots and agents)\label{sub:snnsuitab}}

The Capabilities shown by insects (given their lower neural complexity compared to vertebrates)  to interact and cope with the environment including: exploration, reliable navigation, pattern recognition and interactions with each other, are being considered as key features for implementation in the design of autonomous robots \cite{Helgadottir2013}. Based on the fact that SNNs reproduce to some extent the computational characteristics of biological neural systems, SNNs prove to be a potential computational instrument to achieve the above mentioned features in artificial systems. There is increasing research (e.g., \cite{Liuetal2007}, \cite{Helgadottir2013}, \cite{Cyr2012}, \cite{Wang2014} ) on the use of SNNs to control autonomous systems which exhibit intelligent behaviour in terms of learning and adaptation to the environment. Wang et al. \cite{Wang2014} compares the implementation of SNNs with traditional ANNs autonomous controllers highlighting the following advantages in the SNN approach: (1) Espatio-temporal information is used more efficiently in SNN than in ANNs. (2) The topology of the SNNs controller is much simpler than in ANNs. (3) The (hebbian) training method in the SNNs was easier to implement than in ANNs. SNNs are demonstrating that their application as artificial neural controllers in autonomous systems is not only advantageous in computational terms (when compared with previous connectionists models) but it also allows the implementation of biologically inspired neural systems (e.g., \cite{Helgadottir2013}, \cite{Haenicke2012}, \cite{Hausler2011}) to be used in machines.

\subsection{The Spiking Neural Network (SNN) model\label{sub:snn}}

The SNN model implemented in Netlogo \cite{Wilensky:1999} is a simplified but functional version of integrate and fire neuron models \cite{Maass:1998} \cite{Gerstner:2002} aimed at pedagogical purposes and experimentation with small spiking neural circuits. 
The artificial neuron is implemented as a finite-state machine where the states transitions depend on a variable which represents  the membrane potential of the cell. All the characteristics of the artificial neuron including:  (1) membrane potential, (2) resting potential, (3) spike threshold, (4) excitatory and inhibitory postsynaptic response, (5) exponential decay rate and (6) absolute and refractory periods, are enclosed in two possible states:  open and absolute-refractory.

\begin{figure}[h!]
\centering
\includegraphics[width=3.5in]{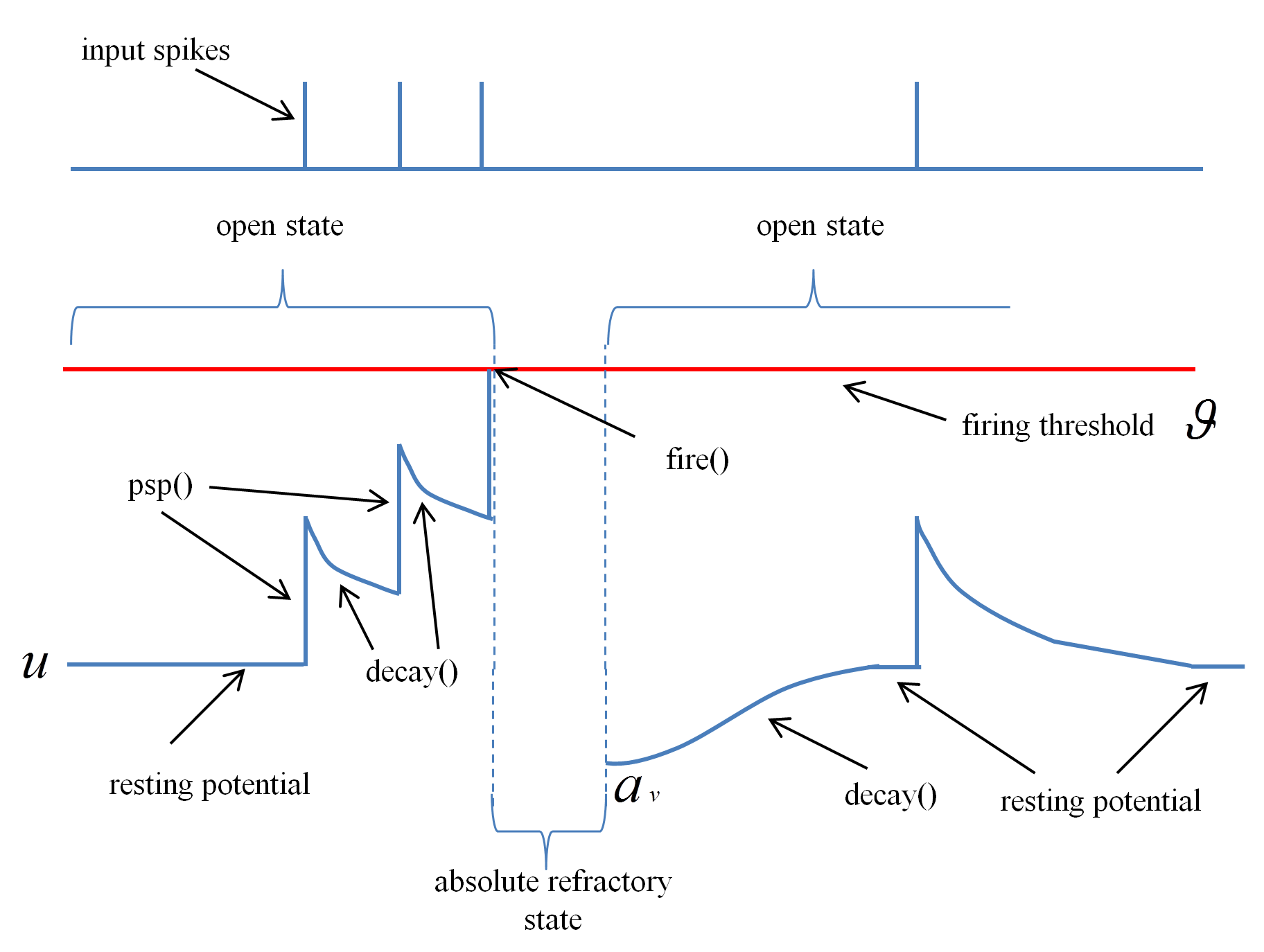}
\caption{Modeling of the membrane potential in the implemented SNN model\label{fig:snn_model}}
\end{figure}

In the open-state the artificial neuron is receptive to input pulses coming from presynaptic neurons. The amplitude of postsynaptic potentials elicited by presynaptic pulses is given by the function $psp()$ (see figure 1). The membrane perturbations reported by $psp()$ are added (excitatory postsynaptic potential EPSP) or subtracted (inhibitory postsynaptic potential IPSP) to the actual value of the membrane potential $u$. If the neuron firing threshold $\vartheta$ is not reached by  $u$, then $u$ begins to decay ($decay()$ function in figure 1) towards a fixed resting potential.  On the other hand, if the membrane potential reaches a set threshold, an action potential or spiking process is initiated. In the used model, when $u$ reaches the firing threshold $\vartheta$, this triggers a state transition from the open to the absolute-refractory state. During the latter, $u$ is set to a fixed refractory potential value $a_v$ and all incoming presynaptic pulses are neglected by $u$. Fig. \ref{fig:snn_model} illustrates the behaviour of the membrane potential in response to incoming presynaptic spikes.

\subsection{The brain for the virtual ant\label{sub:brain}}

\begin{figure}[ht!]
\centering
\includegraphics[width=3.6in]{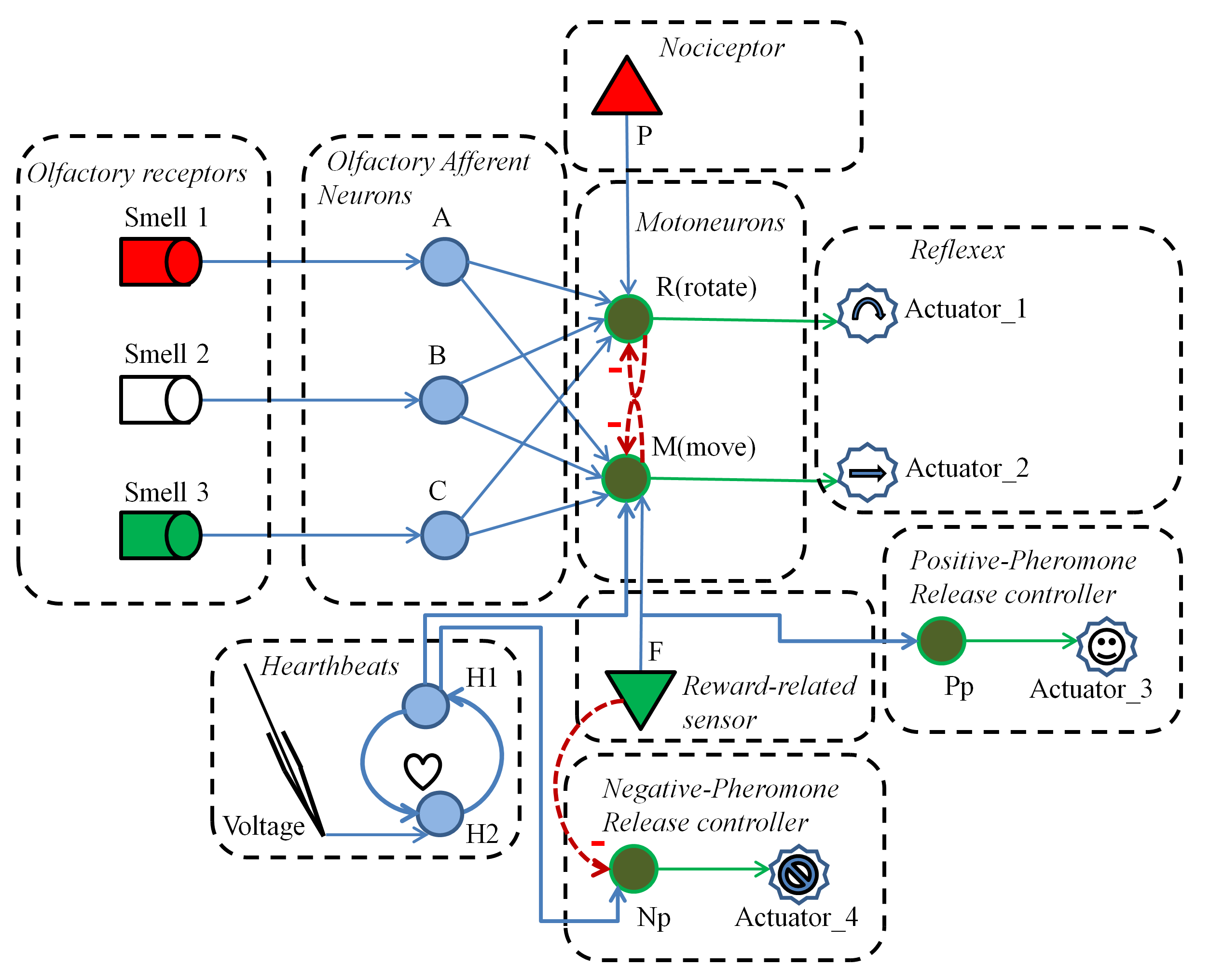}
\caption{Neural circuit controller of the virtual ant\label{fig:complete}}
\end{figure}

\subsubsection{Learning with the Spike Timing Dependent Plasticity (STDP) rule}

In this paper, the STDP model proposed by Gerstner et al. 1996  \cite{Gerstner:1996}  has been implemented as the underlying learning mechanism for the ants neural circuit. In STDP the synaptic efficacy is potentiated or depressed  according to the difference between the arrival time of  incoming pre-synaptic spikes and the time of the action potential triggered at the post-synaptic neuron.

The following formula \cite{Gerstner:1996} describes the weight change of the synapse according to the STDP model for pre-synaptic and post-synaptic neurons $j$ and $i$  respectively:  Here, the arrival time of the pre-synaptic spikes is indicated by $t_j^f$ while $t_i^n$ represents the firing time at the post-synaptic neuron:

\begin{equation}
\Delta w_j = \sum_{j=1}^N{\sum_{n=1}^N{W(t_i^n-t_j^f)}}
\end{equation}

The weight change resulting from the combination of a pre-synaptic spike with a post-synaptic action potential is given by the function $W(\Delta t)$\,\cite{Gerstner:1996}.

\begin{equation}
W(\Delta t) =
\begin{cases}
A_+ \exp{(\Delta t/ \tau_+)},& \text{if } \Delta t < 0\\
-A_- \exp{(-\Delta t/ \tau_-)},& \text{if } \Delta t \geq 0
\end{cases}
\end{equation}

where $\Delta t$ is the time difference between the arriving pre-synaptic spike and the post-synaptic action potential. $A_+$ and $A_-$  determine the amplitude of the weight change when increasing or decreasing respectively. $\tau_+$ and $\tau_-$ determine the reinforce and inhibitory interval or size of the learning window. 

\subsubsection{Topology of the neural circuit}

The neural circuit presented in this work (Fig. \ref{fig:complete})  enables a simulated ant to move in a two dimensional world, learning to identify and avoid noxious stimuli while moving towards perceived rewarding stimuli. At the beginning of the training phase, the ant is not aware of which stimuli are to be avoided or pursued. Learning occurs through  reward-and-punishment classical conditioning \cite{pavlov1927conditioned}. Here the ant learns to associate the information represented by different colours with unconditioned reflex responses. In terms of classical conditioning, learning can be described as the association or pairing of a conditioned or neutral stimulus with an unconditioned stimulus (one that elicits an innate or reflex response). Thus, the neutral or conditioned stimulus acquires the ability to elicit the same response or behaviour produced by the unconditioned stimulus. The pairing of two unrelated stimuli usually occurs by repeatedly presenting the neutral stimulus shortly before the unconditioned stimulus that elicits the innate response. In the context of classical conditioning in animals, the word "shortly" refers to a time interval of a few seconds (or in some cases a couple of minutes). On the other hand, at the cellular level  and in the context of STDP, the association of stimuli encoded as synaptic spikes occurs in short milliseconds intervals \cite{Gerstner:1996}.

\subsubsection{Sensory system}

The neural circuit of the ant is able to process three types of sensorial information: (1) olfactory, (2) pain and (3) pleasant or rewarding sensation.  The olfactory information  is acquired through three olfactory receptors (see figure 2) where each one of them is sensitive to one specific smell represented with a different color (white, red or green). Each olfactory receptor is connected with one afferent neuron which propagates the input pulses towards the Motoneurons. Pain is elicited by a nociceptor whenever the insect collides with a wall or another obstacle. Finally, a rewarding or pleasant sensation is elicited when the insect gets in direct contact with a positive stimulus (i.e. food).  

\subsubsection{Motor system}

The motor system allows the virtual ant to move forward or rotate in one direction according to the reflexive behaviour described below in Fig. \ref{fig:complete}. In order to keep the ant moving even in the absence of external stimuli, the motoneuron M is connected to a neural oscillator sub-circuit composed of two neurons H1 and H2 (Fig. \ref{fig:complete}) performing the function of a pacemaker which sends a periodic pulse to M. The pacemaker is initiated by a pulse from an input neuron which represents an external input current (i.e; intracellular electrode). Fig. \ref{fig:complete} illustrates the complete neural anatomy of the ant.

\subsubsection{Pheromone system}

The positive and negative pheromone systems are controlled by the two neurons Pp and Np respectively. The neuron Pp is activated by the reward sensor F resulting in the release of positive pheromone whenever the ants gets in contact with food (or any other positive stimulus associated with the activation of F). On the other hand, the neuron Np is activated by the summation of pulses coming periodically from the oscillator sub-circuit (neurons H1 and H2). Np works as an energy-consumption counter which fires unless it is inhibited by the reward sensor F. Thus, whenever the ant gets in contact with food the energy counter is reinitialized.

\subsection{The Double Pheromone as the basis for collective intelligence\label{sub:pheromone}}

Traditional models of ant colony systems use mainly a positive pheromone feedback mechanism, meaning that they simulate a single chemical being secreted by the ants and that this chemical acts as an attractor for other ants to follow the pheromone trail. Recent findings on colonies of \textit{monomorium pharaonis} ants show that this species uses a negative pheromone to help repel foragers from unrewarding areas of the landscape\,\cite{robinson:2005aa,Robinson:2007}. This empirical evidence shows that this second chemical acts as a `no entry' signal that ants deposit when they find unrewarding paths. 

The application of a double pheromone mechanism in artificial ant systems has been shown to improve the performance of ant colony optimisation problems, as the use of a negative `no entry' pheromone reduces the exploration space in symetric travelling salesman problems\,\cite{sousa-rodrigues:2014traversing-news,Ramos:2013qy,Rodrigues:2011fj,ramos:2011aa}. 

Following these ideas in the proposed model, ants explore the landscape according to rules established by the internal intelligence (or brain) dictated by the responses of the spiking neural network. The deposition of pheromones is controlled by the occurrence under two situations:

\begin{description}
\item[\textbf{Negative Pheromone deposition:}] \hfill \\
Occurs after a threshold time since food was last found by the ant.
\item[\textbf{Positive Pheromone deposition:}] \hfill \\
Occurs immediately after an ant finds food and the deposition of positive pheromone persists for a parameterized amount of time after that.
\end{description}

\subsection{Implementation in Netlogo\label{sub:netlogo}}
 
In Netlogo there are four types of agents: Turtles, patches, links and the observer \cite{tisuenetlogo:2004}. The virtual ants are represented by turtle agents as well as each neuron in the implemented SNN model. Synapses on the other hand are represented by links. The produced pheromone is represented by patches. All simulated entities including the insect, neurons and synapses have their own variables and functions that can be manipulated using standard Netlogo commands. The Netlogo virtual world consists  of a two dimensional grid of patches where each patch corresponds to a point $(x, y)$ in the plane. In a similar way to the turtles, the patches own a set of primitives which allow the manipulation of their characteristics and also the programming of new functionalities and their interaction with other agents. The visualization of the ants and their environment is done through the Netlogo's world-view interface.

The virtual world of the ant is an ensemble of patches of four different colours, where each one of them is associated with a different type of stimulus. White and Red patches are both used to represent harmful stimulus. Thus, if the ant is positioned on a white or red patch, this will trigger a reaction in the ant's nociceptor (pain sensor) and its corresponding neural pathway (Fig. 2). On the other hand, green patches trigger a reaction in the reward sensor of the ants whenever it is positioned on one of them. Black patches represent empty spaces and do not trigger any sensory information in the ant at all.

\begin{figure}[h!]
\centering
\includegraphics[width=5cm]{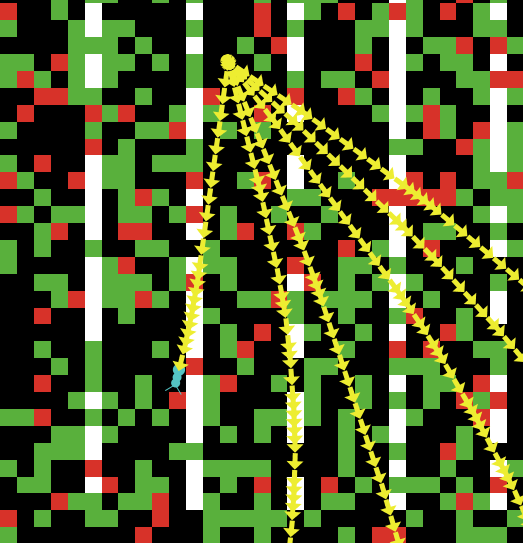}
\caption{Short trajectories at the beginning of the training phase. The ant collides and  escapes  the  world  repeatedly.\label{fig:snnA}}
\end{figure}

\begin{figure}[h!]
\centering
\includegraphics[width=5cm]{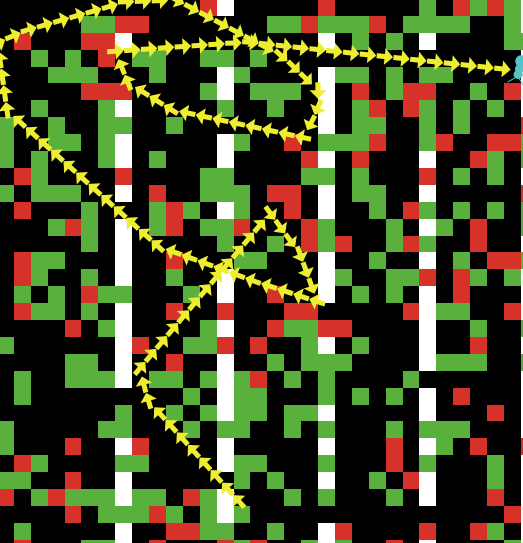}
\caption{Long trajectory shows ant avoiding red and white patches.\label{fig:snnB}}
\end{figure}

At the beginning of the training phase (Fig. \ref{fig:snnA}) the ant moves along the virtual-world colliding indiscriminately with all types of  patches.  the ant is repositioned in its initial coordinates every time it reaches the virtual-world boundaries. As the training phase progresses it can be seen that the trajectories lengthen as the ant learns to associate the red and white patches with harmful stimuli and consequently to avoid them (Fig. \ref{fig:snnB}). 

Once the training phase has been completed, the conditioned aversion to white and red patches is exploited by using the white patches to delimit the boundaries of the virtual world of the ants ( using white as walls ) while the red patches are used to represent the negative pheromone released by the ant.

\section{Results\label{sec:results}}

Figures 5--8 illustrate a sequence of the ants' movements using the double pheromone.

\begin{figure}[h!]
\centering
\includegraphics[width=4cm]{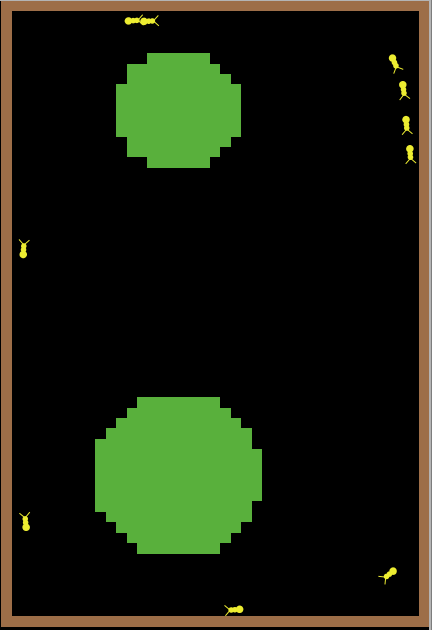}
\caption{No Pheromone.\label{fig:dph0}}
\end{figure}

Fig. \ref{fig:dph0} shows the ants swarm moving through the virtual world. Since the virtual ants can only react to stimuli located directly in front of them , they manage to avoid the obstacles delimiting the virtual world (brown or white patches), however they are not able to detect the food sources located inside the virtual world.

\begin{figure}[h!]
\centering
\includegraphics[width=4cm]{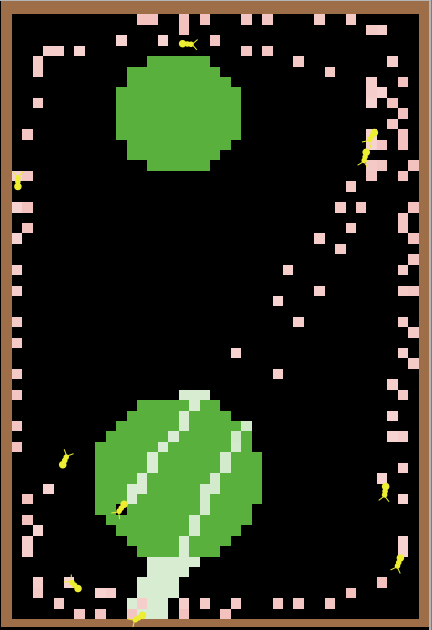}
\caption{After 200 iterations with activated Pheromone.\label{fig:dph1}}
\end{figure}

Fig. \ref{fig:dph1} shows that when the ants start releasing the pheromone their otherwise monotonous trajectories are affected as the pheromone constitutes a new obstacle which is avoided thus creating several random trajectories for the ants.

\begin{figure}[h!]
\centering
\includegraphics[width=4cm]{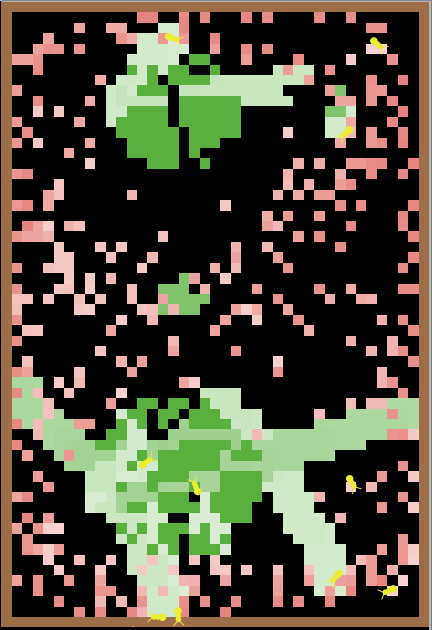}
\caption{After 1000 iterations with activated Pheromone.\label{fig:dph2}}
\end{figure}

\begin{figure}[h!]
\centering
\includegraphics[width=4cm]{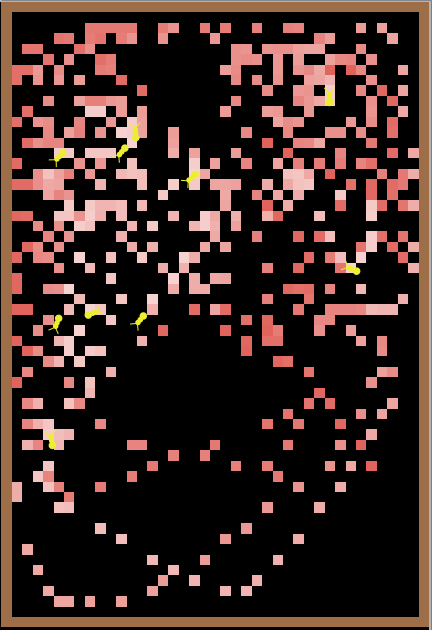}
\caption{After 8000 iterations with activated Pheromone\label{fig:dph3}}
\end{figure}

\begin{figure}[h!]
\centering
\includegraphics[width=9cm]{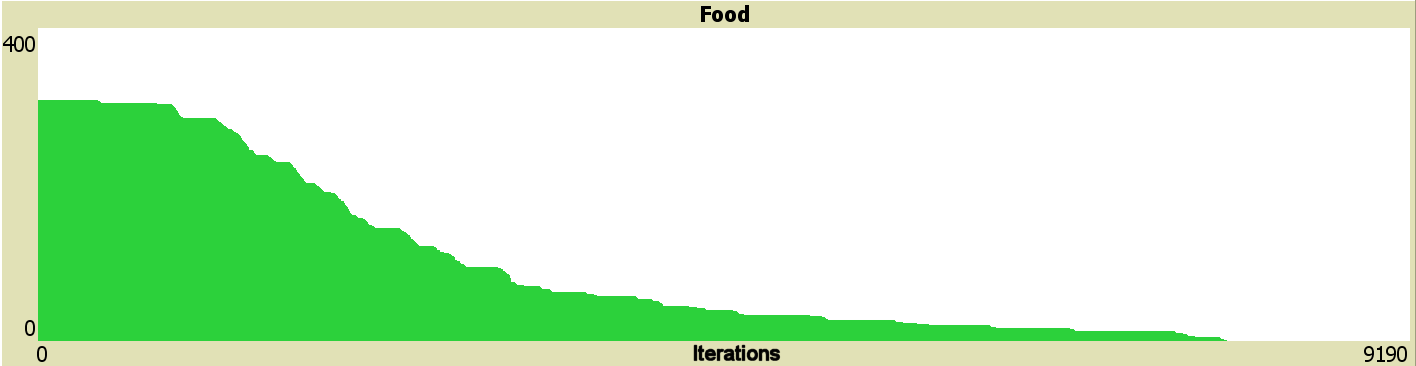}
\caption{Amount of available food during simulation with activated pheromone.\label{fig:dph4}}
\end{figure}

Fig. \ref{fig:dph2} and \ref{fig:dph3} shows that the ants follow new trajectories which allow them to find the food and eat it. At the same time the negative pheromones released, by occupying previously empty spaces in the virtual world  reduce the areas where the ants move to find food (the search space). Fig. \ref{fig:dph3} shows that after several iterations the food sources have been exhausted. The foraging activity is shown in Fig. \ref{fig:dph4} that illustrates the available food over time in the landscape.

\section{Conclusion\label{sec:conclusion}}
Combining internal and external forms of intelligence---or at least forms of individual and societal decision processes---is a challenging problem. It is one task that benefits from building on top of biological findings. In this work an agent based model of an Ant System was presented were both the individual ants and the colonies are organised based on biological principles. 

The model encompasses a combined synergy between internal individual intelligence and external collective intelligence. The former is presented in the form of a Spiking Neural Network while the latter is comprised of a double pheromone based space exploration and \textit{mass recruitment}. 

It was shown how a Spiking Neural Network can be used to endow each ant with the ability to recognise rewarding and harmful stimuli.  
In this paper we demonstrated how associative learning in SNN can be used to allow accurate navigation of virtual ants in a two dimensional environment.  Although the demonstrated association tasks are based on simplified action-reaction (sensor-actuator) relationships, it is possible to extend the neural circuit to associate more complex input patterns (i.e. bitmap images and other sensor arrays) with different types of actuators in order to produce more complex and intelligent behaviour.

It was also shown how the results obtained by the individual ants are enhanced at the collective level by using a double pheromone mechanism. This self-organisation principle allows the collective---even in a small scale model as the one presented---to exhibit emergent features and to solve the problem of foraging by communicating through pheromone deposited in the landscape. This deposition acts as a memory---even if temporary, because of the action of evaporation---that allows the colony to improve their foraging task.
 
The model depicts two forms of intelligence in an easy to use and easy to understand software package that introduces two important paradigms of artificial life to a vast community of scientists.





%
\bibliographystyle{IEEEtran}
\bibliography{references}

\end{document}